\documentclass[11pt]{article}
\usepackage{fullpage,times}
\usepackage{parskip}
\usepackage[utf8]{inputenc} 
\usepackage[T1]{fontenc}    
\usepackage{hyperref}       
\usepackage{url}            
\usepackage{booktabs}       
\usepackage{amsfonts}       
\usepackage{nicefrac}       
\usepackage{microtype}      
\usepackage{xcolor}         
\usepackage{natbib}
\usepackage{graphicx}
\usepackage{subfigure}

\usepackage{lipsum}
\usepackage{bbm}
\usepackage{algorithm}
\usepackage{algorithmic}
\usepackage{amsmath}
\usepackage{multirow}
\usepackage{hhline}

\usepackage{amsmath, amsthm, amssymb, amsfonts, mathtools, graphicx, enumerate}
\usepackage{adjustbox}
\usepackage[utf8]{inputenc} 
\usepackage[T1]{fontenc}    
\usepackage{url}            
\usepackage{booktabs}
\usepackage{nicefrac}       
\usepackage{microtype}      
\usepackage{xspace}
\usepackage{algorithm,algorithmic}
\usepackage{color}
\usepackage{enumitem}
\usepackage{comment}
\usepackage{bm}
\usepackage{apptools}
\usepackage[page, header]{appendix}
\usepackage{titletoc}
\usepackage{array}

\usepackage[scaled=.9]{helvet}

\usepackage{url}
\usepackage{caption}

\usepackage{amsthm}   
\newtheorem{thm}{Theorem}[section]

\newtheorem{pro}[thm]{Proposition}    
\begingroup
    \makeatletter
    \@for\theoremstyle:=definition,remark,plain\do{%
        \expandafter\g@addto@macro\csname th@\theoremstyle\endcsname{%
            \addtolength\thm@preskip\parskip
            }%
        }
\endgroup

\newcommand{\myempty}[1]{}

\newcommand{\R}{\mathbb{R}}

\renewcommand{\R}{\mathbb{R}}
\renewcommand{\P}{\mathcal{P}}

\newcommand{\yrm}[1]{}

 \def\bb#1\ee{\begin{align*}#1\end{align*}}

 \def\bba#1\eea{\begin{align}#1\end{align}}

\makeatletter
\newcommand{\printfnsymbol}[1]{%
  \textsuperscript{\@fnsymbol{#1}}%
}
\makeatother

\title{Efficient Forecasting of Large Scale \\ Hierarchical Time Series via Multilevel Clustering}

\author{
	\small{Xing Han} \\
    ~\small{UT Austin}\\
	\small{\texttt{aaronhan223@utexas.edu}}
	\and
	\small{Tongzheng Ren} \\
	~\small{UT Austin}\\
	\small{\texttt{tongzheng@utexas.edu}} \\
	\and 
	\small{Jing Hu} \\
	~\small{Intuit AI}\\
	\small{\texttt{jing\_hu@intuit.com}} \\
	\and 
	\small{Joydeep Ghosh} \\
	~\small{UT Austin}\\
	\small{\texttt{jghosh@utexas.edu}} \\
	\and
	\small{Nhat Ho} \\
	~\small{UT Austin}\\
	\small{\texttt{minhnhat@utexas.edu}} \\
}

\date{}

\begin{document}

\maketitle

\begin{abstract}
We propose a novel approach to the problem of clustering hierarchically aggregated time-series data, which has remained an understudied problem though it has several commercial applications. We first group time series at each aggregated level, while simultaneously leveraging local and global information.  The proposed method can cluster hierarchical time series (HTS) with different lengths and structures. For common two-level hierarchies, we employ a combined objective for local and global clustering over spaces of discrete probability measures, using Wasserstein distance coupled with Soft-DTW divergence. For multi-level hierarchies, we present a bottom-up procedure that progressively leverages lower-level information for higher-level clustering. Our final goal is to improve both the accuracy and speed of forecasts for a larger number of HTS needed for a real-world application. To attain this goal, each time series is first assigned the forecast for its cluster representative, which can be considered as a ``shrinkage prior'' for the set of time series it represents. Then this base forecast can be quickly fine-tuned to adjust to the specifics of that time series. We empirically show that our method substantially improves performance in terms of both speed and accuracy for large-scale forecasting tasks involving much HTS.
\end{abstract}

\section{Introduction}
Forecasting time-series with hierarchical aggregation constraints is a  problem encountered in many practically important scenarios \citep{hyndman2011optimal, hyndman2016fast, lauderdale2020model, taieb2017coherent, zhao2016multi}. For example, retail sales and inventory records are normally at different granularities such as product categories, store, city and state \citep{makridakis2020m5, seeger2016bayesian}. Generating forecasts for each aggregation level is necessary for developing both high-level and detailed views for marketing insights. Different from multivariate time series forecasting, results for HTS need to be coherent across the given structure. In addition, data at different aggregation levels possess distinct properties w.r.t. sparsity, noise distribution, sampling frequency etc., so utilizing structural information when training forecasting model is helpful for improving final results \citep{han2021simultaneously}. In applications like the e-commerce domain, one HTS typically represents a single user's record, and the number of users is very large. Some users have abundant records while some are limited. Therefore, it is inefficient to build predictive models independently for each HTS. We propose a novel and effective way to address this problem by assigning one model per group of similar HTS as determined by a flexible clustering procedure, followed by fine-tuning of forecasts for each user. In addition to providing superior clustering results for multi-level hierarchies, our approach shows significant improvements in both computational time and accuracy of forecasts when applied to large HTS datasets containing hundreds of thousands of time series.

Time series clustering is an important tool for discovering patterns over sequential data when categorical information is not available. In general, clustering methods are divided into discriminative (distance-based clustering) and generative (model-based clustering) approaches. Discriminative approaches normally define a proper distance measure \citep{muller2007dynamic, petitjean2011global} or construct features \citep{aghabozorgi2015time, paparrizos2015k} that capture temporal information. Generative approaches \citep{zhong2003unified} specify the model type (e.g., HMM) apriori and estimate the parameters using maximum likelihood algorithms. This type of method works better when the underlying modelling assumption is reasonable, but is hampered if the models are misspecified. In addition, generative approaches are more computationally heavy given the maximum likelihood estimation. Deep learning models have also been applied on time series clustering. Popular methods are normally discriminative approaches that first extract useful temporal representations followed by clustering in the embedding space \citep{franceschi2019unsupervised, ma2019learning}. These methods are widely used for clustering time series with different dimensions. For HTS, simply clustering as multivariate time series without considering hierarchy will lead to inferior performance, particularly when the hierarchy is complex. On the other hand, it is difficult to completely respect the hierarchy when clustering, since the data is not easy to partition given the imposed constraints. 

There is little prior work on clustering of multilevel structured data. A pioneering effort \citep{ho2017multilevel} proposed to partition data in both local and global levels and discover latent multilevel structures. This work proposes an optimization objective for two-level clustering based on Wasserstein metrics. Its core idea is to perform global clustering based on a set of local clusters. However, this work mainly applies to discrete and semi-structured data such as annotated images and documents. It cannot be applied on HTS involving continuous and structured data, which has more constraints. A follow-up work \citep{ho2019probabilistic} has extended this to continuous data by assuming that the data at the local level is generated by predefined exponential family distributions. It then performs model-based clustering at both levels. However, model-based clustering for time series is computationally expensive and crucially depends on modeling assumptions. Moreover, both these works are limited to two-level structures, whereas for several HTS applications, given a set of pre-specified features as aggregation variables, it is possible to have a multi-level hierarchy. 

In this paper, we propose an effective clustering method tailored for HTS data. Our method, called \texttt{HTS-Cluster}, is designed for both two-level and multi-level hierarchies. \texttt{HTS-Cluster} employs a combined objective that involves clustering terms from each aggregated level. This formulation uses Wasserstein distance metrics coupled with Soft-DTW divergence \citep{blondel2021differentiable} to cater to variable length series that are grouped together. Overall, \texttt{HTS-Cluster} is an efficient model-free method which can handle HTS with various types of individual components and hierarchies.
\section{Background} \label{sec:back}

\begin{figure}[t]
    \centering
    \setlength{\tabcolsep}{1pt} 
    \renewcommand{\arraystretch}{1} 
    \begin{minipage}{.5\textwidth}
    \includegraphics[width=\textwidth]{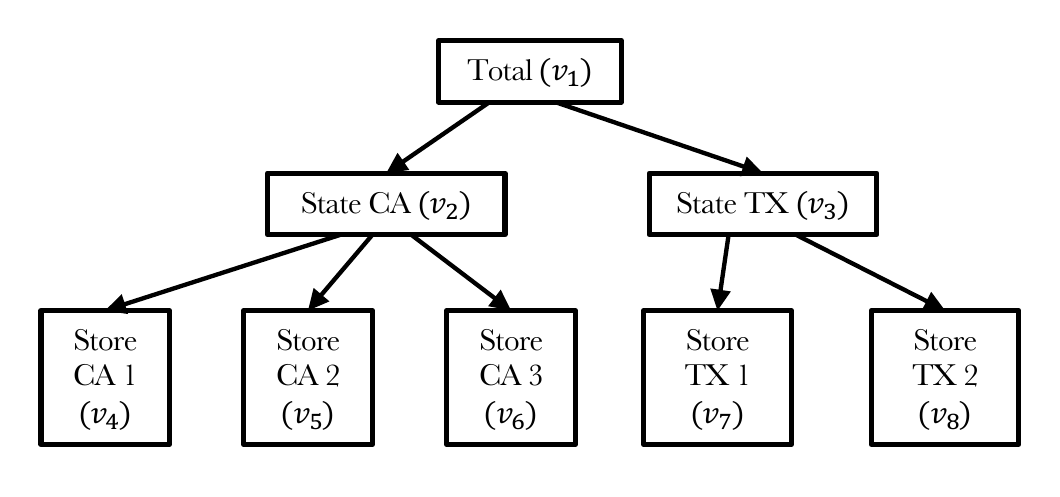}
    \end{minipage}
    \begin{minipage}{.3\textwidth}
    \begin{align*}
    S &= \left[\begin{array}{c}
    S_{\mathrm{aggregated}} \\
    I_5
    \end{array}\right] &=
    \left[\begin{array}{ccccc}
        1&      1&   1&      1&      1 \\
        1&      1&   1&      0&      0 \\
        0&      0&   0&      1&      1 \\
      \hline 
      \multicolumn{5}{c}{\smash{\raisebox{-.5\normalbaselineskip}{$I_5$}}} \\
      & & & &
    \end{array}\right]
\end{align*}
    \end{minipage}
    \caption{Left: an example of HTS \citep{makridakis2020m5} with 5 bottom-level time series and 3-level hierarchical structure. Each vertex ($V = \{v_i\}_{i=1}^8$) represents time series aggregated on different variables related through a domain-specific conceptual hierarchy (e.g., product categories, locations etc). Right: the summation matrix $S \in \{0, 1\}^{8 \times 5}$ that is used to denote the given hierarchy.}
    \label{fig:hierarchical}
\end{figure}

\paragraph{Hierarchical Time Series}
Given the time stamps $t = 1, \dots, T$, let $\textbf{x}_t \in \mathbb{R}^n$ be the value of HTS at time $t$, where $x_{t, i} \in \mathbb{R}$ is the value of the $i^{\mathrm{th}}$ (out of $n$) univariate time series. Figure \ref{fig:hierarchical} shows an example of HTS with three-level structure. We refer to time series at the leaf nodes of the hierarchy as bottom-level time series and the remaining nodes as aggregated-level time series. We split the vector of $\textbf{x}_t$ into $m$ bottom time series and $l$ aggregated time series such that $\textbf{x}_t = [\textbf{a}_t ~~ \textbf{b}_t]^{\top}$ where $\textbf{a}_t \in \mathbb{R}^l$ and $\textbf{b}_t \in \mathbb{R}^m$ with $n = l + m$. The summation matrix $S \in \{0, 1\}^{n\times m}$ satisfy $\textbf{x}_t = S\cdot \textbf{b}_t$, which can later be used to calibrate forecasting results to be aligned with given hierarchical structure. For notational simplicity, we will omit the time stamp of each series in the following discussion.

\paragraph{Dynamic Time Warping}
Dynamic time warping (DTW) \citep{sakoe1978dynamic, muller2007dynamic} is a popular method for computing the optimal alignment between two time series with arbitrary lengths. Given $\mathbf{X}$ and $\mathbf{Y}$ of length $T_1$ and $T_2$, DTW computes the $T_1\times T_2$ pairwise distance matrix between each time stamp and solves a dynamic program (DP) using Bellman's recursion in $\mathcal{O}(T_1 \cdot T_2)$ time. DTW discrepancy can be used to describe average similarity within a set of time series \citep{petitjean2011global}. However, DTW is not a differentiable metric given its DP recursion nature. To address this issue, \citet{cuturi2017soft} proposed soft-DTW by smoothing the $\texttt{min}$ operation using \texttt{log-sum-exp} trick. Specifically, assume $\mathbf{A}\in \{0, 1\}^{T_1 \times T_2}$ is the alignment matrix between two time series and $\mathbf{C} \in \R^{T_1 \times T_2}$ is the cost matrix, the formulation of Soft-DTW can be written as 
\begin{equation}
    \mathrm{SDTW}_{\gamma}(\mathbf{C}(\mathbf{X}, \mathbf{Y})) = \underset{\mathbf{A}\in \mathcal{A}(T_1, T_2)}{\min}^{\gamma} \langle \mathbf{A}, \mathbf{C} \rangle = -\gamma \log \sum_{\mathbf{A} \in \mathcal{A}(T_1, T_2)} \exp (-\langle\mathbf{A, C}\rangle / \gamma),
\label{eq:sdtw}
\end{equation}
where $\gamma > 0$ is a parameter that controls the trade-off between approximation and smoothness and $\mathcal{A}(T_1, T_2)$ is the collection of all possible alignments between two time series.
Soft-DTW is differentiable w.r.t. all of its variables and can be used for a variety of tasks such as averaging, clustering and prediction of time series. However, soft-DTW also has several drawbacks. \citet{blondel2021differentiable} recently showed that soft-DTW is not a valid divergence given its minimum is not achieved when two time series are equal; also, the value of soft-DTW is not always non-negative. \citet{blondel2021differentiable} proposed soft-DTW divergence, which can address these two issues and achieves better performance in relevant tasks. This divergence, denoted as $D_{\gamma}^C$, can be written as
\begin{equation}
    D_{\gamma}^C (\mathbf{X}, \mathbf{Y}) :=  \mathrm{SDTW}_{\gamma}(C(\mathbf{X}, \mathbf{Y})) - \frac{1}{2} \mathrm{SDTW}_{\gamma} (C(\mathbf{X}, \mathbf{X})) - \frac{1}{2} \mathrm{SDTW}_{\gamma} (C(\mathbf{Y}, \mathbf{Y})).
    \label{eq:sdtwd}
\end{equation}
Our method will incorporate the soft-DTW divergence as a base distance measure for variable length sequences, and use it as a differentiable loss during the clustering procedure.

\paragraph{Wasserstein Distances}
For any given subset $\Theta \subset \R^d$, let $\mathcal{P}(\Theta)$ denote the space of Borel probability measures on $\Theta$. The Wasserstein space of order $r$ of probability measures on $\Theta$ is defined as $\mathcal{P}_r (\Theta) = \left\{G\in \mathcal{P}(\Theta): \int\|x\|^r dG(x) <\infty\right\}$, where $\|\cdot\|$ denotes Euclidean metric in $\R^d$. For any element $P, Q$ in $\mathcal{P}_r(\Theta)$, the $r$-Wasserstein distance $W_r$ between $P$ and $Q$ is defined as~
\begin{equation}
    W_r(P, Q) = \left( \underset{\pi \in \Pi(P, Q)}{\inf} \int_{\Theta^2} \|x - y\|^r ~d\pi(x, y) \right)^{1/r},
    \label{eq:wass}
\end{equation}
where $\Pi(P, Q)$ contains all the joint (coupling) distributions whose marginal are $P$ and $Q$, and the coupling $\pi$ that achieves the minimum of Eq \eqref{eq:wass} is called the transportation plan. In other words, $W_r(P, Q)$ is the optimal cost of moving mass from $P$ to $Q$, which is proportional to $r$-power of Euclidean distance in $\Theta$. Furthermore, by recursion of concepts, define $\mathcal{P}_r(\mathcal{P}_r(\Theta))$ as the space of Borel measures on $\mathcal{P}_r(\Theta)$, then for any $P', Q' \in \mathcal{P}_r(\mathcal{P}_r(\Theta))$, we have 
\begin{equation}
    W_r^{(2)}(P', Q') = \left(\underset{\pi \in \Pi(P', Q')}{\inf} \int_{\mathcal{P}_r(\Theta)^2} W_r^r(P, Q) ~d\pi(P, Q)\right)^{1/r}. \nonumber
\end{equation}
Similarly, the cost of moving unit mass in its space of support $\mathcal{P}_r(\Theta)$ is proportional to the $r$-power of the $W_r$ distance in $\mathcal{P}_r(\Theta)$. The Wasserstein distance can be thought as a special case of Wasserstein barycenter problem. The computation of Wasserstein distance and Wasserstein barycenter has been studied by many prior works \citep{agueh2011barycenters, alvarez2016fixed, benamou2015iterative, solomon2015convolutional, lin2019efficient, Lin-2020-Revisiting, lin2019complexity}, where~\citet{cuturi2014fast} proposed an efficient algorithm for finding its local solutions. The well-known $K$-means clustering algorithm can also be viewed as solving a Wasserstein means problem~\citep{pollard1982quantization, ho2017multilevel}. 
\section{Hierarchical Time Series Clustering} \label{sec:method}

Denote $x_{j, i}$ as the $i^{\mathrm{th}}$ univariate time series of the $j^{\mathrm{th}}$ HTS, where $1 \leq j \leq N, ~ 1\leq i \leq n_j$. Assume the index $i$ of each univariate series is given by the level-order traversal of the hierarchical tree from left to right at each level. We will use $a_{j, i}$ and $b_{j, i}$ for the corresponding aggregated and bottom-level series. We start with clustering two-level time series and then extend to multiple levels.

\subsection{Two-Level Time Series Clustering} \label{sec:sdtw}
\begin{figure}[t]
    \centering
    \includegraphics[width=\textwidth]{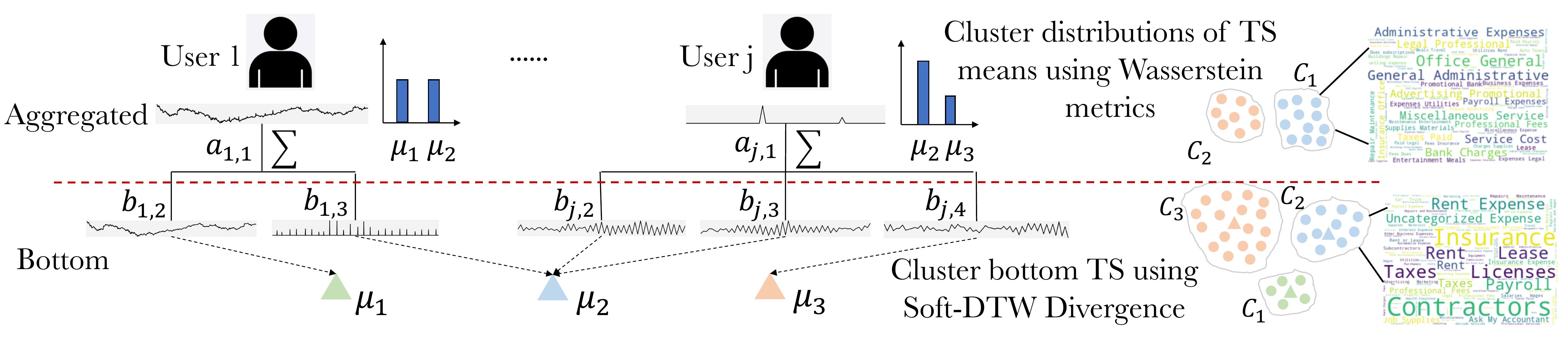}
    \caption{Leveraging local clustering results for hierarchical time series clustering. We improve  performance at aggregated level by clustering empirical distributions over bottom-level means.}
    \label{fig:demo}
\end{figure}

Given that data at the same aggregated level possess similar properties, we perform time series clustering in a level-wise fashion. Define $W_{\mathrm{sdtw}}$ as a new Wasserstein distance measure by replacing the Euclidean distance in Eq \eqref{eq:wass} with soft-DTW divergence $D_{\gamma}^C$ defined in Eq \eqref{eq:sdtwd}. For any $j = 1, \dots, N$, we denote the empirical measure of all bottom-level series as $P_{N'} = \frac{1}{N'}\sum_{j=1}^N\sum_{i=1}^{n_j} \delta_{b_{j,i}}$, where $N' \geq N$ given that each HTS has at least one bottom-level series. For local (bottom-level) clustering, assume that at most $k_2$ clusters can be obtained, we perform $K$-means that can be viewed as finding a finite discrete measure $G =  \sum_{k=1}^{k_2} u_k \delta_{\mu_k} \in \mathcal{O}_{k_2}(\Theta)$ that minimizes $W_{\mathrm{sdtw}}(G, P_{N'})$, where $\mu_k \in \mathbb{R}^T$ is the ``cluster mean'' time series to be optimized as the support of the finite discrete measure $G$ and $u \in \Delta_{k_2}$. Although this approach can be extended to any aggregated level, such method cannot leverage the connections with adjacent levels. Figure \ref{fig:demo} givens an example on how summation causes lose of information in time series: it is less likely to obtain reasonable results by simply clustering data at aggregated level. Given the loss of information from the aggregation operation, it can be helpful to perform top-level clustering with the help of bottom-level series. 

\paragraph{Problem Formulation} 
A direct solution is to replace each top level series by a large feature vector obtained by concatenating all bottom level series, but this will introduce redundancy and require large training datasets due to the induced high dimensionality. Instead, we propose to leverage local information by utilizing bottom-level clustering results. For the $j^{\mathrm{th}}$ HTS, we denote $\mathcal{F}_j(i)$ as the set that contains all descendant indices of its $i^{\mathrm{th}}$ time series, and assume each top level series $a_{j, 1}$ is aggregated from the bottom level series $\{b_{j,i}\}_{i \in \mathcal{F}_j(1)}$, we first cluster all bottom level series $\{b_{j,i}\}_{j\in [N], i\in \mathcal{F}_j(1)}$ into $k_2$ clusters $\{C_k\}_{k\in [k_2]}$ with $C_k$ centered at $\mu_k$. We then assign the following probability measure to each $a_{j, 1}$: 
\begin{equation}
    a_{j, 1} = \frac{1}{|\mathcal{F}_j(1)|} \sum_{i\in \mathcal{F}_j(1)} \sum_{k\in [k_2]} \mathbf{1}_{\mu_k} \mathbf{1}_{b_{j, i} \in C_k},
    \label{eq:hts}
\end{equation}
i.e. we represent the top level time series as an empirical distribution of $\{\mu_k\}_{k\in [k_2]}$, where the weight of each $\mu_k$ is determined by the number of $b_{j, i}$ that belongs to cluster $C_k$. This formulation represents the top-level time series using finite number of bottom-level clusters, which reduces computation time from concatenating bottom-level time series while simultaneously leverages local information. We now define the objective function for jointly optimizing both local and global clusters as follows
\begin{equation}
    \underset{\substack{G\in\mathcal{O}_{k_2}(\Theta),\\ \mathcal{H} \in \mathcal{O}_{k_1}(\mathcal{P}_r(\Theta))}}{\inf}  W_{\mathrm{sdtw}}(G, P_{N'}) + W_{\mathrm{sdtw}}^{(2)}(\mathcal{H}, \frac{1}{N} \sum_{j=1}^N \delta_{a_{j, 1}}),
    \label{eq:jo}
\end{equation}
where $k_1$ is the number of clusters at the top level, and $\mathcal{H}$ is the distribution over top-level cluster centroids. Denote $\Delta_k = \left\{ w\in \R^k: w_i \geq 0, \sum_{i=1}^k w_i=1\right\}$ as the probability simplex for any $k \geq 1$. Similar to each top-level time series $a_{j, 1}$, the supports of $\mathcal{H}$ are also finite discrete measures. Specifically, $\mathcal{H} = \sum_{k=1}^{k_1} v_k\delta_{\nu_k} \in \mathcal{P}_r(\mathcal{P}_r(\Theta))$, where $v \in \Delta_{k_1}$, $\nu_k \in \mathcal{O}_{\bar{n}}(\Theta)$ and $\bar{n} = \frac{1}{N}\sum_{j=1}^N \mathcal{F}_j(1)$. Eq \eqref{eq:jo} is our formulation for the two-level hierarchical time series clustering problem, where the first term $W_{\mathrm{sdtw}}(G, P_{N'})$ is Wasserstein distance defined in the space of measures $\mathcal{P}_r(\Theta)$ and the second term is defined in $\mathcal{P}_r(\mathcal{P}_r(\Theta))$.

\paragraph{Leveraging Local and Global Information}
We now propose an algorithm to solve Eq \eqref{eq:jo} by performing alternating updates of time series clusters at each aggregated level. At each iteration, we first perform one step of cluster assignment and centering for $\{b_{j,i}\}_{j\in [N], i\in \mathcal{F}_j(1)}$, which returns a set of cluster means $\{\mu_k\}_{k\in [k_2]}$. They are then used to construct the probability measure $\{a_{j, 1}\}_{j\in[N]}$ by Eq \eqref{eq:hts}. The iteration is finished by another step of cluster assignment and centering over $\{a_{j, 1}\}_{j\in[N]}$. This process is repeated until the cluster assignment of both top and bottom levels are stable. Overall, the alternating update can better utilize structural information during hierarchical time series clustering: cluster means obtained from $\{b_{j,i}\}_{j\in [N], i\in \mathcal{F}_j(1)}$ can serve as features when clustering $\{a_{j, 1}\}_{j\in[N]}$; simultaneously, top-level clustering results can also be used as global information to calibrate bottom-level clustering (e.g., assigning boundary time series given clusters are not well-separated). Empirically, this method is effective in clustering simple two-level HTS.

\begin{algorithm}[t]
\small
\caption{Bottom-Up Clustering for Multilevel HTS}
\label{alg:htsc}
\begin{algorithmic}[1]
\STATE \textbf{Input}: $\mathsf{L}$, total aggregation level; $x_{1:N}^l$, collection of the $l^{\mathrm{th}}$ level series from \#1 to \#$N$ HTS
\STATE \textbf{Initialize}: cluster assignment: $\{C^l_k\}_{k \in [\mathsf{k}_l], l \in [\mathsf{L}]}$
\WHILE{not converged}
\STATE compute cluster mean by $\mu_k^{\mathsf{L}} = \mathop{\arg\min}_{\mu\in \mathbb{R}^T} \frac{1}{|C_k^{\mathsf{L}}|}\sum_{i\in C_k^{\mathsf{L}}} D_{\gamma}^C(x_{1:N}^{\mathsf{L}}[i], \mu),  ~ \forall k \in [\mathsf{k}_{\mathsf{L}}].$
\STATE obtain cluster assignment by $C^{\mathsf{L}}_k = \{i: D_{\gamma}^C(x_{1:N}^{\mathsf{L}}[i], \mu_k^{\mathsf{L}}) = \underset{k\in [\mathsf{k}_{\mathsf{L}}]}{\min} D_{\gamma}^C  (x_{1:N}^{\mathsf{L}}[i], \mu_k^{\mathsf{L}}) \}, ~ \forall k \in [\mathsf{k}_{\mathsf{L}}].$
\ENDWHILE
\FOR{$l = [\mathsf{L}-1, \mathsf{L}-2, \dots, 1]$}
\STATE def measure for data at $l^{\mathrm{th}}$ level: $\textbf{x}_j^l[n] = \frac{1}{|\mathcal{F}_j(n)|} \sum_{i\in \mathcal{F}_j(n)} \sum_{k\in [\mathsf{k}_{l+1}]} \mathbf{1}_{\nu_k^{l+1}} \mathbf{1}_{x_j^{l+1}[i] \in C_k^{l+1}}, \forall j \in [N]$.
\WHILE{not converged}
\STATE compute Wasserstein barycenter by $\nu_k^{l} = \mathop{\arg\min}_{\nu \in \P_r(\Theta)} \sum_{i\in C_k^{l}} \lambda_i W_{\mathrm{sdtw}}(\textbf{x}_{1:N}^l[i], \nu), ~ \forall k \in [\mathsf{k}_{l}].$
\STATE obtain cluster assignment by $C^{l}_k = \{i: W_{\mathrm{sdtw}} (\textbf{x}_{1:N}^l[i], \nu_k^{l}) = \underset{k\in [\mathsf{k}_{l}]}{\min} W_{\mathrm{sdtw}} (\textbf{x}_{1:N}^l[i], \nu_k^{l}) \}, ~ \forall k \in [\mathsf{k}_{l}].$
\ENDWHILE
\STATE update cluster mean by $\nu_k^{l} = \mathop{\arg\min}_{\nu\in \mathbb{R}^T} \frac{1}{|C_k^{l}|}\sum_{i\in C_k^{l}} D_{\gamma}^C(x_{1:N}^{l}[i], \nu),  ~ \forall k \in [\mathsf{k}_{l}].$
\ENDFOR
\end{algorithmic}
\end{algorithm}

\subsection{Multi-Level Time Series Clustering}
For HTS with multiple levels, we employ a ``bottom-up'' clustering procedure that recursively uses lower level information for higher-level clustering, till the root is reached. Assume we are given $N$ number of HTS with $\mathsf{L}$ levels; denote $x_{1:N}^l$ as collection of the $l^{\mathrm{th}}$ level time series from the first $N$ HTS and $\textbf{x}_{1:N}^l$ as the replacement of corresponding time series represented by lower level clusters. We formulate the objective function for clustering multilevel time series as
\begin{equation}
    \underset{\substack{G\in\mathcal{O}_{\mathsf{k_L}}(\Theta),\\ \mathcal{H}^l \in \mathcal{O}_{\mathsf{k}_l}(\mathcal{P}_r(\Theta))}}{\inf}  W_{\mathrm{sdtw}}(G, P_{N'}) + \sum_{l=1}^{\mathsf{L}-1} W_{\mathrm{sdtw}}^{(2)}(\mathcal{H}^l, \frac{1}{\mathcal{G}(l)} \sum_{j=1}^{\mathcal{G}(l)} \delta_{\mathbf{x}_{1:N}^l[j]}),
    \label{eq:multilevel}
\end{equation}
where $\mathcal{G}(l)$ is the total number of time series at level $l$ among $N$ HTS. Similarly, we have $\mathcal{H}^l = \sum_{k=1}^{\mathsf{k}_l} v_k\delta_{\nu_k^l}$, $v\in\Delta_{\mathsf{k}_l}$ and $\nu_k^l \in \mathcal{O}_{\bar{n}_l}(\Theta)$. Algorithm \ref{alg:htsc} shows the full procedure of the bottom-up clustering for HTS with arbitrary levels, while its core steps also apply on two-level HTS clustering. Specifically, step 4 and 5 are the centering and cluster assignment steps for clustering time series at the $\mathsf{L}^{\mathrm{th}}$ (bottom) level, which uses soft-DTW divergence as a distance measure between each input pair. After the clustering result at the $\mathsf{L}^{\mathrm{th}}$ level is obtained, its cluster indices and means are used to construct probability measures for each time series at the $\mathsf{L} - 1$ level (step 8). Meanwhile, we still have access to the original time series in aggregated levels. Step 10 and 11 perform clustering in the space of probability measures. To efficiently compute Wasserstein barycenter, we optimize the support of barycenter $\nu_k^l$, which is featured as the free-support method studied by \cite{cuturi2014fast}. For cluster assignment, we compute Wasserstein distance between pairs of probability measures in the space of soft-DTW divergence. Both the assignment and centering steps utilizes information from lower aggregation level, where these steps are repeated until the cluster assignments are stable. We then use the results of cluster assignment to compute cluster means of the original time series at that level, which are used as supports of probability measure to represent time series in the next aggregation level, until the clustering procedure for all levels are finished.

\begin{pro} \label{thm:converge}
Algorithm \ref{alg:htsc} monotonically decreases the objective function Eq \eqref{eq:multilevel}.
\end{pro}

\paragraph{Computational Efficiency}
Overall, the HTS clustering framework we proposed is an efficient approach while leveraging the predefined hierarchical information. Compared with model-based clustering, it waives extensive computation of HMM parameters and the data assumptions. In addition, the centering steps (4 and 10) of each level can be optimized using gradient-based methods. Notice that, solving step 10 is much faster than solving step 4 even if same amount of time series are clustered: in step 4, one needs to compute the soft-DTW divergence between cluster means $\{\mu_k^{\mathsf{L}}\}_{k\in \mathsf{k_L}}$ and all time series $x_{1:N}^{\mathsf{L}}$ at that level; while in step 10, one just need to compute the soft-DTW divergence between the supports of multiple distributions, i.e., constructing the transportation cost with size $\mathsf{k_L} \times \mathsf{k_L}$, which can be obtained beforehand. Furthermore, the number of time series at a higher aggregation level is evidently smaller than lower levels. Therefore, the clustering time will be progressively reduced as we proceed to higher aggregation levels. As we shall see later, the speeds gains for the associated forecasting problem are even greater.


\subsection{On the Choice of the Number of Clusters} 
Determining the number of clusters $K$ is a well-known model selection problem. We note that clustering is an intermediate step in our final goal of large scale forecasts of a set of HTS, and using a larger value of $K$
can lead to some extra computation but does not detract from this goal. 
Therefore 
we initialize each $\mathsf{k}_l$ to be a conservatively chosen large, and apply a post-processing method after clustering to eliminate "redundant" clusters by the merge  and remove operations. Specifically, taking the bottom-level time series as an example, given the obtained cluster means $\{\mu_k^{\mathsf{L}}\}_{k\in [\mathsf{k_L}]}$ and the cluster indices $\{C_k^{\mathsf{L}}\}_{k\in[\mathsf{k_L}]}$, we merge cluster $C_i^{\mathsf{L}}$ and $C_j^{\mathsf{L}}$, if $D_{\gamma}^C(\mu_i^{\mathsf{L}}, \mu_j^{\mathsf{L}}) < \epsilon_M$, where $\epsilon_M$ is some pre-defined parameter to determine if the two clusters are close enough, and we remove the cluster $C_i^{\mathsf{L}}$ if $|C_i^{\mathsf{L}}| \leq \epsilon_R$, i.e.,  if the cluster is too small. After we finish merging and/or removing clusters as needed, we can perform an additional fine-tuning on the remaining clusters. Finally, for time series in aggregated levels, we just need to replace $D_{\gamma}^C$ with the Wasserstein distance $W_{\mathrm{sdtw}}$. 

\subsection{Clustering for Hierarchical Time Series Forecasting}
Forecasts for individual HTS can ``borrow strength'' from the forecasts of nearest cluster means at each aggregated level. Specifically, we first perform forecast for bottom-level and aggregated level cluster-mean time series $\{\mu_k^{\mathsf{L}}\}_{k\in [k_{\mathsf{L}}]}$ and $\{\nu_k^l\}_{k\in [k_l], l\in [\mathsf{L} - 1]}$. The forecast of each univariate time series can be represented as the weighted combination of forecasts of corresponding cluster means at that level. We define the weight between time series $i$ and cluster mean $j$ at level $l$ as 
\begin{equation}
    w_{i,j}^l = \frac{1}{\sum_{k=1}^{k_l} \left(\frac{D_{\gamma}^C(x_i^l, \nu_j^l)}{D_{\gamma}^C(x_i^l, \nu_k^l)} \right)^{\frac{2}{m - 1}}}, \quad m \in (1, \infty),
    \label{eq:fuzzy}
\end{equation}
where the closer $x_i^l$ to a certain cluster mean, the higher its weight is. This equation is well-known in fuzzy clustering \citep{yang1993survey}, where a data point can belong to more than one cluster, and $m$ is the parameter that controls how fuzzy the cluster assignments is.
One can use post-reconciliation methods such as \citet{wickramasuriya2019optimal} to calibrate the results for individual time series forecasts.

\section{Experiments} \label{sec:exp}
We evaluate our method (HTS-Cluster) in multiple applications over different datasets. Overall, our experiments include: (1) clustering time series with multilevel structures (Section~\ref{sec:cluster}); (2) facilitating time series forecasting with the help of clusters (Section~\ref{sec:forecast}).

\subsection{Time Series Clustering} \label{sec:cluster}

\begin{table}[t]
\centering
\renewcommand\arraystretch{1.2}
\scalebox{0.77}{
\begin{tabular}{c|c|ccc|ccc}
\specialrule{1.5pt}{1pt}{1pt}
\multirow{2}{*}{ Method$\backslash$Metric} & \multirow{2}{*}{Time} & \multicolumn{3}{c}{Global} & \multicolumn{3}{c}{Local} \\ \hhline{~|~|---|---}
& & NMI & AMI & ARI & NMI & AMI & ARI \\ \hline
\texttt{DTCR} & 132 & 0.325$\pm$\texttt{.012} & 0.257$\pm$\texttt{.023} & 0.21$\pm$\texttt{.011} & 0.392$\pm$\texttt{.014} & 0.313$\pm$\texttt{.006} & 0.284$\pm$\texttt{.009} \\
\texttt{Soft-DTW} & 67 & 0.412$\pm$\texttt{.009} & 0.326$\pm$\texttt{.019} & 0.277$\pm$\texttt{.008} & 0.411$\pm$\texttt{.022} & 0.342$\pm$\texttt{.009} & 0.304$\pm$\texttt{.014} \\
\texttt{Concat} & 186 & 0.436$\pm$\texttt{.015} & 0.342$\pm$\texttt{.014} & \textbf{0.314}$\pm$\texttt{.016} & 0.411$\pm$\texttt{.022} & 0.342$\pm$\texttt{.009} & 0.304$\pm$\texttt{.014} \\
\textbf{HTS-Cluster} & \textbf{37} & \textbf{0.455}$\pm$\texttt{.018} & \textbf{0.354}$\pm$\texttt{.015} & 0.302$\pm$\texttt{.013} & \textbf{0.424}$\pm$\texttt{.018} & \textbf{0.366}$\pm$\texttt{.013} & \textbf{0.321}$\pm$\texttt{.018} \\ \hhline{|=|=|===|===|}
\texttt{DTCR} & 72 & 0.065$\pm$\texttt{.002} & 0.015$\pm$\texttt{.001} & 0.008$\pm$\texttt{.002} & 0.105$\pm$\texttt{.011} & 0.059$\pm$\texttt{.002} & 0.054$\pm$\texttt{.003} \\
\texttt{Soft-DTW} & 49 & 0.119$\pm$\texttt{.005} & 0.043$\pm$\texttt{.003} & 0.027$\pm$\texttt{.003} & 0.126$\pm$\texttt{.008} & \textbf{0.082}$\pm$\texttt{.006} & 0.061$\pm$\texttt{.005} \\
\texttt{Concat} & 174 & \textbf{0.135}$\pm$\texttt{.004} & 0.073$\pm$\texttt{.007} & \textbf{0.045}$\pm$\texttt{.006} & 0.126$\pm$\texttt{.008} & \textbf{0.082}$\pm$\texttt{.006} & 0.061$\pm$\texttt{.005} \\
\textbf{HTS-Cluster} & \textbf{34} & 0.134$\pm$\texttt{.005} & \textbf{0.075}$\pm$\texttt{.005} & 0.041$\pm$\texttt{.004} & \textbf{0.128}$\pm$\texttt{.014} & 0.064$\pm$\texttt{.005} & \textbf{0.065}$\pm$\texttt{.002} \\
\specialrule{1.5pt}{1pt}{1pt}
\end{tabular}}
\caption{Level-wise clustering results on HTS with two aggregated levels. The upper part shows results on simulated data. The lower part gives results on real-world financial record data using a weak proxy for the (unknown) cluster labels, hence all the numbers are low.}
\label{tab:sim_nmi}
\end{table}

\paragraph{Two-Level HTS} We first conduct experiments on \textit{synthetic data} using ARMA simulations, to provide a feel for the setting and the results attainable. We generate a simple HTS with two levels: one parent node with 4 children vertices, i.e., for the $j^{\mathrm{th}}$ hierarchy $X_j = \{x_i\}_{i=1}^5, x_1 = \sum_{i=2}^5 x_i$. The length of each $X$ is different, ranging  from $80$ to $300$. We use the following simulation function for each time series $x_{1:T}$ 
$$
x_t = 0.75 x_{t - 1} - 0.25 x_{t - 2} + 0.65 \varepsilon_{t-1} + 0.35 \varepsilon_{t-2} + \varepsilon_t + c,
$$
where $\varepsilon_t$ is a white noise error term at time $t$, and $c$ is an offset that is used to separate different clusters. We simulate 4 clusters,   each having $30$ HTS as members. Additionally, the evaluation is also performed on a real-world HTS dataset from a large cooperation. This dataset contains $12,000$ users' electronic record of expenses in different categories. The bottom-level time series are summed across all categories to obtain total expense records. Each user possesses an HTS but the length of records varies from user to user. 

\begin{table}[t]
\centering
\renewcommand\arraystretch{1.2}
\scalebox{0.8}{
\begin{tabular}{c|c|c|c|c|c|c|c|c|c}
\specialrule{1.5pt}{1pt}{1pt}
\multirow{2}{*}{ Level } & \multirow{2}{*}{ Metric } & \multicolumn{4}{c}{Simulation} & \multicolumn{4}{c}{Financial Record} \\ \hhline{~|~|----|----}
& & \texttt{DTCR} & \texttt{Soft-DTW} & \texttt{Concat} & \textbf{HTS-Cluster} & \texttt{DTCR} & \texttt{Soft-DTW} & \texttt{Concat} & \textbf{HTS-Cluster} \\ \hhline{--|----|----}
\multirow{3}{*}{ 1 } & NMI & 0.28 & 0.313 & 0.342 & \textbf{0.356} & 0.037 & 0.124 & \textbf{0.156} & 0.154 \\
& AMI & 0.243 & 0.277 & 0.301 & \textbf{0.322} & 0.021 & 0.079 & \textbf{0.112} & 0.106 \\
& ARI & 0.221 & 0.265 & 0.285 & \textbf{0.304} & 0.009 & 0.056 & \textbf{0.094} & 0.092 \\ \hline
\multirow{3}{*}{ 2 } & NMI & 0.298 & 0.317 & 0.357 & \textbf{0.375} & 0.056 & 0.116 & 0.147 & \textbf{0.152} \\
& AMI & 0.271 & 0.282 & 0.314 & \textbf{0.346} & 0.034 & 0.087 & 0.115 & \textbf{0.121} \\
& ARI & 0.236 & 0.259 & 0.302 & \textbf{0.317} & 0.016 & 0.034 & 0.083 & \textbf{0.092} \\ \hline
\multirow{3}{*}{ 3 } & NMI & 0.272 & 0.324 & 0.364 & \textbf{0.372} & 0.055 & 0.134 & 0.163 & \textbf{0.172} \\
& AMI & 0.234 & 0.295 & 0.322 & \textbf{0.33} & 0.028 & 0.098 & 0.132 & \textbf{0.141} \\
& ARI & 0.217 & 0.268 & 0.307 & \textbf{0.309} & 0.012 & 0.057 & 0.106 & \textbf{0.113} \\ \hline
\multirow{3}{*}{ 4 } & NMI & 0.303 & \textbf{0.369} & \textbf{0.369} & \textbf{0.369} & 0.076 & \textbf{0.136} & \textbf{0.136} & \textbf{0.136} \\
& AMI & 0.275 & \textbf{0.341} & \textbf{0.341} & \textbf{0.341} & 0.043 & \textbf{0.102} & \textbf{0.102} & \textbf{0.102} \\
& ARI & 0.264 & \textbf{0.316} & \textbf{0.316} & \textbf{0.316} & 0.026 & \textbf{0.061} & \textbf{0.061} & \textbf{0.061} \\
\specialrule{1.5pt}{1pt}{1pt}
\end{tabular}}
\caption{Level-wise clustering results on HTS with multiple aggregated levels. On the left are results on simulated data while the
right shows results on real-world user financial record data. Since cluster labels are not available for financial data, scores obtained from a weak proxy are lower as expected.}
\label{tab:multilevel_nmi}
\end{table}

\vspace{-2ex}
\paragraph{Multilevel HTS} We also test our method on HTS with multiple aggregated levels. It is simple to extend simulated two-level HTS to multiple levels by modifying the summation matrix $S$. Evaluation is also performed on a large, real-world dataset that contains HTS with $\geq 3$ levels. Each HTS represents expense records of a small business, where the bottom level (or the lowest two levels) time series are user defined accounts (or sub-accounts), which are then aggregated by different tax purposes to obtain the middle level time series. The top level time series are total expenses aggregated from middle level: it includes the overall information of the business. The dataset contains 18,568 HTS with 222,989 bottom level time series in total.

\begin{figure*}[t!]
    \begin{minipage}{\textwidth}
    \centering
    \begin{tabular}{@{\hspace{-7ex}} c @{\hspace{-9ex}} c @{\hspace{-9ex}} c}
        \begin{tabular}{c}
        \includegraphics[width=.42\textwidth]{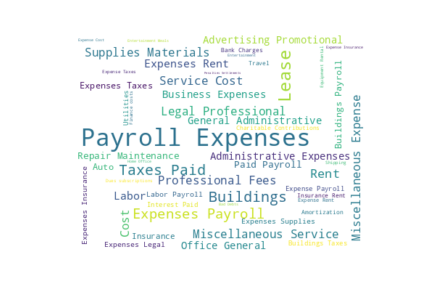}
        \end{tabular} &
        \begin{tabular}{c}
        \includegraphics[width=.42\textwidth]{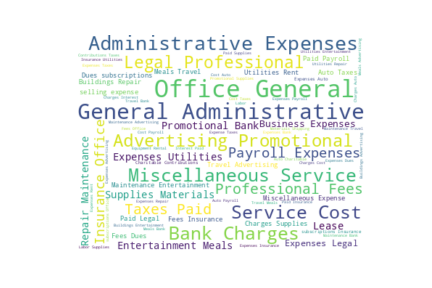}
        \end{tabular} &
        \begin{tabular}{c}
        \includegraphics[width=.42\textwidth]{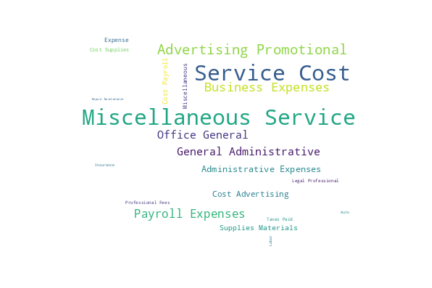}
        \end{tabular} \\
        \end{tabular}
    \end{minipage}
    \vspace{-2ex}
    \caption{Word cloud visualization of time series meta data at the aggregated level of tax code.}
    \label{fig:meta}
\end{figure*}

\begin{table}[t]
\centering
\renewcommand\arraystretch{1.2}
\scalebox{0.93}{
\begin{tabular}{c|ccccc|ccccc}
\specialrule{1.5pt}{1pt}{1pt}
Dataset & \multicolumn{5}{c}{Wiki} & \multicolumn{5}{c}{M5} \\ \hline
Levels & 1 & 2 & 3 & 4 & 5 & 8 & 9 & 10 & 11 & 12 \\ \hline
LSTNet & 76.36 & 76.89 & 79.65 & 81.13 & \textbf{86.22} & 63.74 & 69.43 & 73.35 & \textbf{76.46} & \textbf{82.36} \\
LSTNet-Cluster & \textbf{76.33} & \textbf{76.56} & \textbf{77.68} & \textbf{78.07} & 95.16 & \textbf{62.48} & \textbf{69.14} & \textbf{71.11} & 76.52 & 98.78 \\ \hline
DeepAR & \textbf{73.98} & 74.54 & 77.42 & 79.12 & \textbf{84.77} & 59.36 & 67.18 & \textbf{72.04} & 76.41 & \textbf{80.24} \\
DeepAR-Cluster & 74.21 & \textbf{74.37} & \textbf{77.36} & \textbf{77.56} & 89.67 & \textbf{58.74} & \textbf{65.46} & 74.39 & \textbf{75.04} & 90.49 \\
\specialrule{1.5pt}{1pt}{1pt}
\end{tabular}}
\caption{HTS-Cluster can be used to improve HTS forecasting, when a large number of forecasts need to be done. Results are measured by mean absolute scaled error $(\mathrm{MASE}^{\downarrow})$ using two multivariate time series models. Both Wiki and M5 possess a single hierarchy with many time series; we cluster ``sub-trees'' at the bottom 2 levels (out of 5) of Wiki and the bottom 3 levels (out of 12; we only show levels 8 to 12) of M5 to reduce total number of time series to be modeled.}
\label{tab:hts_forecast}
\end{table}

\vspace{-2ex}
\paragraph{Experiment Baselines} Our baselines for evaluating HTS-Cluster include the recent state-of-the-art method DTCR \citep{ma2019learning}, which employs an encoder-decoder structure integrated with a fake sample generation strategy. \cite{ma2019learning} shows that DTCR can learn better temporal representations for time series data that improves clustering performance. Here, we implement DTCR to treat hierarchical time series as regular multivariate time series data. In addition, we also implemented independent level-wise clustering using Soft-DTW divergence (\texttt{Soft-DTW}) i.e., without local information and clustering aggregated level data via simply concatenating lower level time series (\texttt{concat}). We use three prevalent methods for clustering evaluation: Normalized Mutual Information (NMI) \citep{schutze2008introduction}, Adjusted Mutual Information (AMI) \citep{hubert1985comparing}, and Adjusted Rand Index (ARI) \citep{vinh2010information}. 
\vspace{-2ex}
\paragraph{Clustering Results} We conducted 10 experiments, with different random seeds, on both simulated and real datasets. As shown in Table \ref{tab:sim_nmi} (upper), for the synthetic two-level HTS, our method is superior to the baseline methods in both clustering performance and computational efficiency. Specifically, in terms of clustering performance, level-wise clustering approaches are better than DTCR, at both global (aggregated) and local (bottom) levels, since separating information from different granularities can improve the partitioning of data. As for computation time, DTCR training consists of two stages: it first learns temporal representations and then performs $K$-means clustering. This results in longer computation time compared with HTS-Cluster. For level-wise approaches, clustering using Soft-DTW divergence and simple concatenation yield the same results at the bottom level, but concatenating bottom level data provides better results at the top level since aggregation causes lose of information. Finally, the  alternating updates using the global and local cluster formulations of Eq \eqref{eq:jo}, leads to improved performance due to  leveraging both local and global information. Specifically, the top level time series are represented by empirical distributions over bottom-level cluster means, and the cluster means at the top level can be obtained more efficiently via fast computation of Wasserstein barycenter. Based on user specified domain knowledge or constraints, we utilize the global cluster assignment to calibrate local time series that are far from the nearest cluster center. This procedure improves both local and global clustering results while simultaneously reducing total computation time.

HTS-Cluster also demonstrates improved performance over baseline methods on multilevel HTS. As shown in Table \ref{tab:multilevel_nmi}, all methods are evaluated on HTS datasets with 4 aggregation levels, where level 1 is the top level and 4 is the bottom level. Here, HTS-Cluster employs the bottom-up procedure of Algorithm \ref{alg:htsc}, where the clustering results from lower level are leveraged for upper level clustering, until the root is reached. Therefore, the level-wise clustering methods (\texttt{Soft-DTW, Concat}, and HTS-Cluster) share the same results at the bottom level. At aggregated levels, HTS-Cluster consistently outperforms \texttt{DTCR} and \texttt{Soft-DTW} with the help of local information, and achieves competitive performance with \texttt{Concat} at a much smaller computational cost. 

For the financial data, there are no cluster labels. So we use the "business type", which is included in the meta data of each HTS, as a weak  "ground truth" label for clustering. Not surprisingly, the metrics for all
the methods are low as a result (Table 1 bottom), and the utility of HTS-Cluster really emerges when we examine the downstream forecasting results later on.
For now, to show that the clusters are still meaningful, we visualize the 
 HTS meta data at the level of tax code using our method (Figure \ref{fig:meta}).
and see that HTS-Cluster does create meaningful partitions for HTS by taking into account features from local time series. Finally, we monitor the level-wise clustering time of each method. As shown in Table \ref{tab:htsf} (left), HTS-Cluster provides the most efficient way of clustering aggregated level time series among all level-wise clustering approaches.

\begin{table}[t]
\renewcommand\arraystretch{2}
\hspace{-2ex}
	\begin{minipage}{0.4\linewidth}
		\centering
		\includegraphics[width=\textwidth]{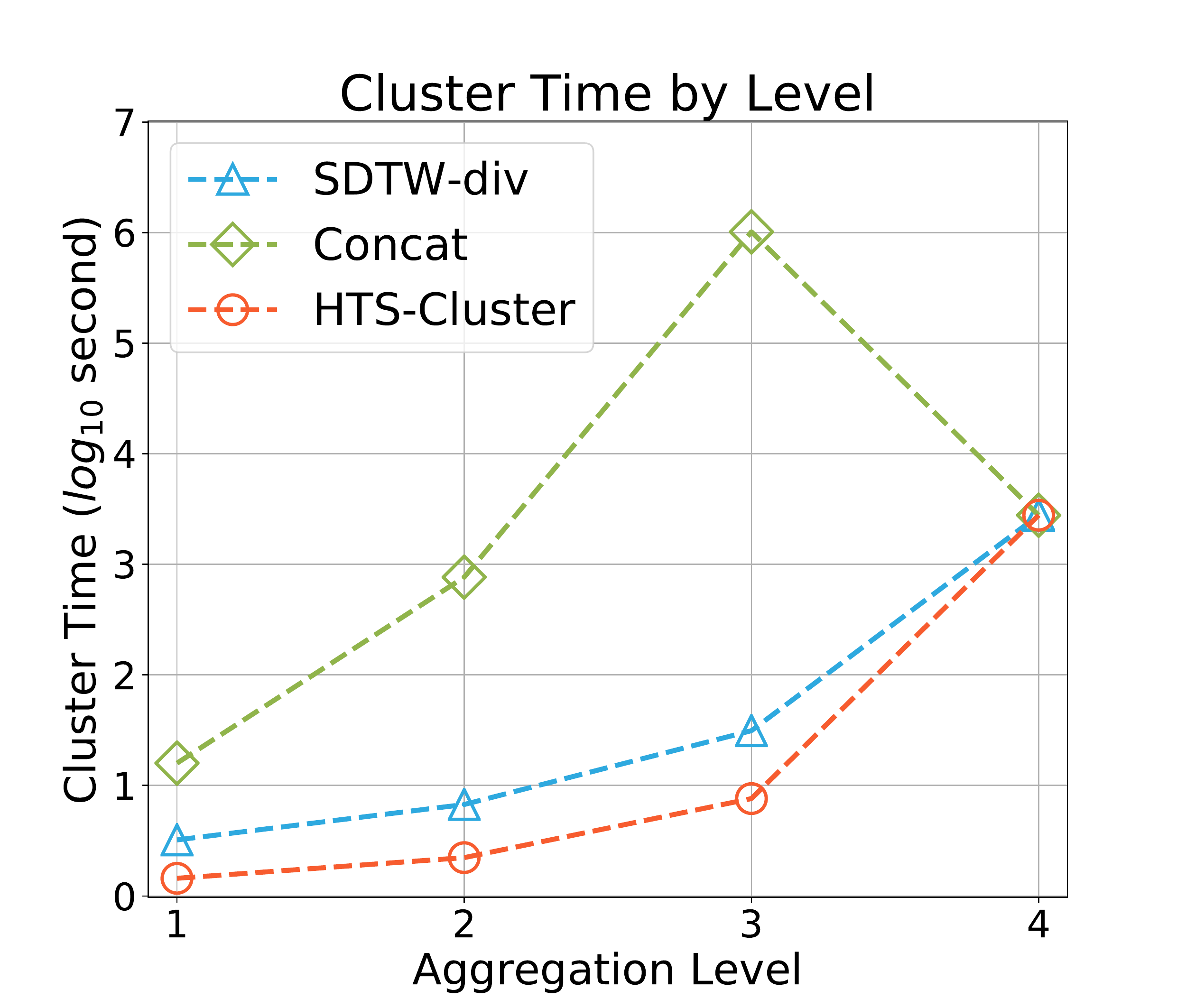} 
	\end{minipage}\hfill
	\begin{minipage}{0.6\linewidth}
		\centering
		\vspace{5.5ex}
        \scalebox{0.82}{
        \begin{tabular}{c|c|c|c|c|c}
        \specialrule{1.5pt}{1pt}{1pt}
        Method $\backslash$ Level & 1 & 2 & 3 & 4 & Total Time \\ \hhline{|=|=|=|=|=|=|}
        W/o cluster & 62.39 & 76.26 & 78.25 & 84.14 & 1 \\ \hline
        DTCR & 82.35 & 96.09 & 104.85 & 104.33 & 0.39 \\
        Soft-DTW & 78.61 & 93.04 & 93.12 & 96.76 & 0.27 \\
        Concat & 74.24 & 84.65 & 83.73 & 96.76 & 0.57 \\
        HTS-Cluster & 72.99 & 80.07 & 85.29 & 96.76 & 0.16 \\
        \specialrule{1.5pt}{1pt}{1pt}
        \end{tabular}} \vspace{4.3ex}
	\end{minipage}
	\vspace{-0.5ex}\caption{Left: level-wise computation time of different clustering approaches. Right: forecasting massive HTS with the help of clustering, results are measured by $\mathrm{MASE}^{\downarrow}$ and relative computing time. All results are averaged across 10 random runs on 4-level simulated HTS.}
    \label{tab:htsf}
\end{table}
\vspace{-2ex}
\subsection{Hierarchical Time Series Forecasting} \label{sec:forecast}
\paragraph{Case 1: Forecast Single HTS with Complex Structure}
Many public HTS datasets consist of a single hierarchy that includes a large number of time series; this is a common situation when HTS in a certain application has many categorical variables to be aggregated. Forecasting a large number of correlated time series requires extensive computation for global models or parameter tuning for local models. HTS-Cluster provides an efficient way of modeling HTS with complex hierarchy. For the bottom $k$ levels that have large numbers of time series, one just needs to forecast their cluster means obtained from clustering ``sub-trees'' at these levels using HTS-Cluster. The forecasts of each time series at the bottom $k$ levels can be ``reconstructed'' using soft combination of cluster means in Eq \eqref{eq:fuzzy}. We test our method using two popular multivariate forecasting models: DeepAR \citep{salinas2020deepar} and LSTNet \citep{lai2018modeling} and two public available HTS datasets: Wiki and M5. As shown in table \ref{tab:hts_forecast}, this strategy achieves competitive performance (measured by mean absolute scaled error (MASE) \citep{hyndman2006another}) with less computation compared with original methods: it could also improve aggregated levels that clustering has not been applied to. 

\vspace{-2ex}
\paragraph{Case 2: Forecast Massive HTS with Simple Structures}
Similarly, we forecast cluster means obtained from each level of HTS, and then use Eq \eqref{eq:fuzzy} to obtain prediction for each HTS. To ensure forecasts are consistent w.r.t. the hierarchical structure, we apply the post-reconciliation method from \cite{wickramasuriya2019optimal} to the obtained forecasts of each HTS. Table \ref{tab:htsf} (right) compares different clustering approaches with building independent model for each HTS without clustering in terms of forecasting accuracy and total computation time. From the results,
HTS-Cluster can greatly reduce the overall computation time without compromising forecasting accuracy.



\section{Conclusion} \label{sec:conclusion}
In this paper, we address an important but under-studied problem for clustering time series with hierarchical structures. Given that time series at different aggregated levels possess distinct properties, regular clustering methods for multivariate time series are not ideal. We introduce a new clustering procedure for hierarchical time series such that when clustering is conducted at the same aggregated level it simultaneously utilizes clustering results from an adjacent level. In each clustering iteration, both local and global information are leveraged.  Our proposed method show improved clustering performance in both simulated and real-world hierarchical time series, and proves  to be an effective solution when a large number of hierarchical time series forecasting needs to be done as a downstream task. For future work, we plan to extend this framework to model-based clustering for hierarchical time series data with some known statistical properties.

\bibliographystyle{bib_style}
\bibliography{reference}

\newpage
\begin{center}
{\bf \huge{Supplementary Material}}
\end{center}
\appendix
\section{Proof of Proposition 3.1}
\label{sec:proof}
\paragraph{Proposition 3.1} 
Algorithm 1 monotonically decreases the objective function Eq (6).
\begin{proof}
To show that Algorithm 1 monotonically decreases Eq (6), we show that the loss function of HTS clustering at each level is guaranteed to decrease monotonically in each iteration for the cluster assignment (line 5 or 11 in Algorithm 1) step and the centering (line 4 or 10) step until convergence. Define the loss function of HTS clustering at the aggregated level $l$ as 
\begin{equation}
    L(\mu) = \sum_{i=1}^N \underset{j \in [k_l]}{\min} D_{\gamma}^C (x_i, \mu_j).
    \label{eq:obj_func}
\end{equation}
Based on the definition of Soft-DTW divergence \citep{blondel2021differentiable}, Eq \eqref{eq:obj_func} is non-negative, then the algorithm will eventually converge when the loss function at each level reaches its (local) minimum. Denote $z = (z_1, \dots, z_N)$ as the cluster assignments for the $N$ time series at level $l$, we now decompose the cluster assignment step and centering step.

\textit{Assignment step:} ~~ The loss function at the aggregated level $l$ can be written as 
\begin{equation}
    L(\mu, z) = \sum_{i=1}^N D_{\gamma}^C (x_i, \mu_{z_i}).
\end{equation}
Consider a time series $x_i$, let $z_i$ be the assignment from the previous iteration and $z_i^*$ be the new assignment obtained as
\begin{equation}
    z_i^* \in \underset{j \in [k_l]}{\arg\min} ~D_{\gamma}^C (x_i, \mu_j).
    \label{eq:assignment}
\end{equation}
Let $z^*$ denote the new cluster assignments for all $N$ time series, then the change of loss function after the assignment step is given by
\begin{equation}
    L(\mu, z^*) - L(\mu, z) = \sum_{i=1}^N \left[D_{\gamma}^C (x_i, \mu_{z_i^*}) - D_{\gamma}^C (x_i, \mu_{z_i})\right] \leq 0.
\end{equation}
The inequality holds by how $z_i^*$ is determined in Eq \eqref{eq:assignment}.

\textit{Centering step:} ~~ The loss function at the aggregated level $l$ can also be written as 
\begin{equation}
    L(\mu, z) = \sum_{j=1}^{k_l} \left[\sum_{i:z_i=j} D_{\gamma}^C (x_i, \mu_j) \right].
\end{equation}
Let $\mu_j \in \mathbb{R}^T$ be the time series cluster mean from the previous iteration and $\mu_j^*$ be the new cluster mean obtained as 
\begin{equation}
    \mu_j^* = \underset{\mu \in \mathbb{R}^T}{\arg\min} \frac{1}{|\{i: z_i = j\}|}\sum_{i: z_i = j} D_{\gamma}^C(x_i, \mu).
    \label{eq:center}
\end{equation}
Let $\mu^*$ denote the new time series cluster means for all $k_l$ clusters, then the change of loss function after the centering step is given by 
\begin{equation}
    L(\mu^*, z) - L(\mu, z) = \sum_{j=1}^{k_l} \left[\sum_{i: z_i = j} D_{\gamma}^C(x_i, \mu_j^*) - \sum_{i: z_i = j} D_{\gamma}^C(x_i, \mu_j) \right] \leq 0.
\end{equation}
This inequality holds since the update of $\mu_j^*$ in Eq \eqref{eq:center} minimizes this quantity. Therefore, at any aggregated level $l$, its objective function Eq \eqref{eq:obj_func} is monotonically decreased at each iteration of Algorithm 1 and so does the objective function Eq (6) for clustering multilevel HTS.
\end{proof} 

\section{Wasserstein Formulation for K-Means Clustering}
\label{sec:Wasserstein_Kmeans}
\paragraph{Wasserstein Barycenter}
Given a set of Borel probability distributions $\{P_i\}_{i=1}^M \in \mathcal{P}_2(\Theta)$ for $M\geq 1$, their Wasserstein barycenter $\overline{P}_{M, \lambda}$ is defined as 
\begin{equation}
    \overline{P}_{M, \lambda} = \arg\min_{P \in \mathcal{P}_2(\Theta)} \sum_{i=1}^M \lambda_i W_2^2 (P, P_i),
    \label{eq:barycenter}
\end{equation}
where $\lambda \in \Delta_M$ is the weight of $\{P_i\}_{i=1}^M$. \citep{cuturi2014fast} proposed an efficient algorithm for finding its local solutions, given $\{P_i\}_{i=1}^M$ are discrete measures over finite number of supports. In step 10 of Algorithm 1, we essentially follow the procedure of free-support method in computing the Wasserstein barycenters as proposed by \citet{cuturi2014fast}.

\paragraph{$K$-Means Clustering}
$K$-means follows the EM procedure by iterating between computing cluster centroid and reassignment until convergence. Assume we have an unlabeled dataset $\{X_i\}_{i=1}^N$ drawn from the sample space $\Theta$. Given an $L_2$ distance measure $D(\cdot, \cdot): \Theta\times\Theta\to \mathbb{R}$ and the number of cluster $k$, $K$-means clustering solves the optimization problem by finding the set $S$ containing at most $k$ cluster centroids \citep{graf2007foundations, pollard1982quantization}: $\min_{S:|S| = k} \frac{1}{N}\sum_{i=1}^N \min_{S_j\in S} D(X_i, S_j).$ Denote $P_N = \frac{1}{N} \sum_{i=1}^N \delta_{X_i}$ as the empirical distribution of the data, the $K$-means formulation can also be interpreted as~\citep{pollard1982quantization, ho2017multilevel} 

\begin{equation}
    \underset{G \in \mathcal{O}_k(\Theta)}{\inf} W_2^2(G, P_N) = 
    \underset{G \in \mathcal{O}_k(\Theta)}{\inf} \inf_{\pi\in \Pi(G, P_N)}\int D (x, y)~ d \pi (x, y),
    \label{eq:ot}
\end{equation}
where $G$ is the empirical distribution of cluster centroids, $\mathcal{O}_k(\Theta)$ is the set of Borel distribution that supports on at most $k$ discrete points in $\Theta$. The Wasserstein formulation of clustering time series data is obtained by replacing $D(x, y)$ in Eq \eqref{eq:ot} with Soft-DTW divergence.

\section{Implementation Details}
\label{sec:implementation_details}

\paragraph{Experiment Settings}
Our experiment is performed on an Ubuntu server with 4 NVIDIA GeForce 3090 GPUs. The implementation of Soft-DTW divergence is publicly available\footnote[1]{https://github.com/google-research/soft-dtw-divergences} with Apache-2.0 license. For the HTS forecasting experiment, we used DeepAR, LSTNet, and dynamic linear model (DLM)\footnote[2]{https://pydlm.github.io/} as the set of forecasting models and results are averaged across each model. We used 1 as the forecasting recursive step and 0.6/0.2/0.2 as the train/validation/test split ratio.


\paragraph{Public HTS Data}
\textit{M5 Competition}\footnote[3]{https://mofc.unic.ac.cy/wp-content/uploads/2020/03/M5-Competitors-Guide-Final-10-March-2020.docx} involves the unit sales of various products ranging from January 2011 to June 2016 in Walmart. It involves the unit sales of 3,049 products, classified into 3 product categories (Hobbies, Foods, and Household) and 7 product departments, where these categories are disaggregated. The products are sold across ten stores in three states (CA, TX, and WI).

\textit{Wikipedia Page View}\footnote[4]{https://www.kaggle.com/c/web-traffic-time-series-forecasting} contains daily views of 145k various Wikipedia articles ranging from July 2015 to December 2016. We sample 150 bottom-level series from 145k series and aggregate to obtain upper-level series. The aggregation features include the type of agent, type of access, and country codes. We then obtain a 5-level hierarchical structure with 150 bottom series.

\paragraph{Financial Data}
This dataset is collected by a large financial software company and forecasts are performed to provide financial insights to each user. We have been granted permission to publish the desensitized results but the full data will not be made public currently considering the user privacy.

\end{document}